\documentclass{article}

\PassOptionsToPackage{numbers, compress}{natbib}
\usepackage[preprint]{neurips_2026}

\usepackage[utf8]{inputenc} 
\usepackage[T1]{fontenc}    
\usepackage{hyperref}       
\usepackage{url}            
\usepackage{booktabs}       
\usepackage{amsfonts}       
\usepackage{nicefrac}       
\usepackage{microtype}      
\usepackage{xcolor}         
\usepackage[pdftex]{graphicx}

\title{Neural Scaling Universality: If Exponents Are Fixed, Time to Understand Coefficients}

\author{%
  Yizhou Liu and Jeff Gore \\
  Massachusetts Institute of Technology\\
  Cambridge, MA 02139 \\
  \texttt{\{liuyz, gore\}@mit.edu} \\
}

\begin{document}

\maketitle

\vskip 0.1in
\begin{abstract}
  Neural scaling laws describe how pre-training loss decays as power laws with training time, model size, and compute. This position paper argues that the exponents of these power laws are fixed by generic mechanisms: a one-third time scaling due to the strong nonlinearity of Softmax, an inverse width scaling due to representational superposition, and an inverse depth scaling due to ensemble averaging of Transformer layers. These mechanisms are robust to a wide range of data structures and architectural details, placing current large language models in a universality class with fixed exponents. The coefficients, however, are expected to be sensitive to data and architecture details, and directly determine practical quantities such as the optimal model shape and the compute-optimal frontier. We therefore argue that understanding the coefficients is the key to near-term performance improvements, and that a closer examination of the current universality class may reveal pathways to better universality classes.
\end{abstract}
\vskip 0.4in

\section{Introduction}

Neural scaling laws, which describe pre-training loss decay with increasing model size, dataset size, and compute as power laws \cite{hestness2017deep, kaplan2020scaling, rae2021scaling, hoffmann2022chinchilla}, are among the key observations that led to the success of today's large language models (LLMs). They describe and predict LLM performance \cite{brown2020gpt3,openai2023gpt4} and have shaped modern training recipes \cite{hoffmann2022chinchilla}. However, the scaling laws remain largely empirical, leaving hope for further speedups and concerns about generalizability at larger scales.

Extensive efforts have therefore been devoted to exploring the mechanistic origins of neural scaling laws. The mainstream view, summarized from a wide range of frameworks \cite{spigler2020asymptotic, hutter2021learning, maloney2022solvable, sharma2022neural, michaud2023quantization, bahri2024explaining, bordelon2025feature, bordelon2025theory}, attributes the power-law loss scaling to some power-law structures in the data. The intuition is that there are various features, skills, or tasks to be learned. If these entities have power-law importance or frequency distributions, and the model learns the more important or frequent ones first, the loss will also decay as power laws whose exponents sensitively depend on the data power laws.

\begin{figure}
  \centering
  \includegraphics{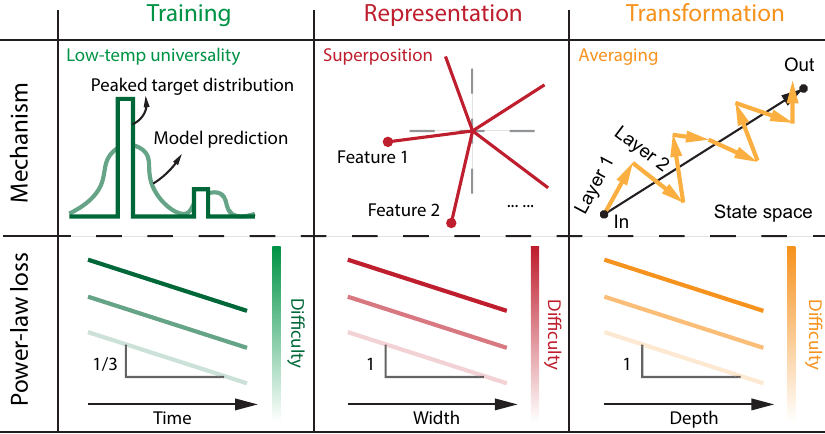}
  \caption{Loss can be decomposed into three leading terms, one due to imperfect training, one due to limited representation, and one due to restricted transformation power. For imperfect training, loss decays largely as a power law with exponent $1/3$ due to the strong nonlinearity of Softmax in reaching peaked distributions, which is related to low-temperature expansion and universality \cite{liu2026universal}. The final value includes a term limited by representation, which is inversely proportional to width, due to geometric interference of more features than dimensions, i.e., superposition \cite{liu2025superposition}. Another part is from limited transformation from input to output, which is inversely proportional to depth, due to averaging of errors across layers with similar functions \cite{liu2026inverse}. Data structures, e.g., a wide range of feature frequency distributions, affecting the difficulty of learning, primarily change the coefficients \cite{liu2025superposition} rather than the exponents.}
  \label{fig:abstract}
\end{figure}

Recently, a new view has arisen, arguing that the power-law loss can emerge without power laws in data \cite{liu2026universal,liu2025superposition,liu2026inverse,liuneural,barkeshli2026origin}, whose exponents are determined by generic mechanisms robust to data details. 
A series of three papers \cite{liu2026universal,liu2025superposition,liu2026inverse} (see Figure~\ref{fig:abstract}) identified (i) a one-third time scaling due to Softmax nonlinearity in learning peaked distributions (e.g., next-token distributions and attention distributions) \cite{liu2026universal}, (ii) an inverse width scaling due to geometric interference of more features than available dimensions, i.e., superposition \cite{liu2025superposition}, and (iii) an inverse depth scaling due to ensemble averaging of errors across Transformer layers \cite{liu2026inverse}. The time scaling can be transferred to dataset size scaling because large-scale pre-training is close to online, one-epoch training \cite{liu2026universal}. Width and depth scaling laws lead to a one-third model size scaling at the optimal aspect ratio \cite{liuneural}, which is determined by the coefficients of the width and depth scaling laws. The optimal token-to-parameter ratio or compute-optimal frontier is also determined by the coefficients, at which the loss follows a one-sixth compute scaling \cite{liuneural}. The power-law exponents are fixed by mechanisms that depend only on high-level architectural and data properties, such as the usage of Softmax and the low entropy of target distributions, neither of which is likely to change. In this framework, a wide range of detailed data structures, e.g., feature frequency distributions, mainly affect the coefficients \cite{liu2025superposition} rather than the exponents. The coefficients therefore warrant close attention, as they directly affect the optimal model shape and compute-optimal frontier and are likely to shift with modified architectures and training strategies.

Based on the three papers \cite{liu2026universal,liu2025superposition,liu2026inverse}, we propose a hypothesis, \textbf{neural scaling universality}: a universality class of neural scaling laws is defined by the set of fixed exponents, where varying architecture or data details changes the coefficients of the power laws but not their exponents. Although more work needs to be done to further verify this hypothesis, evidence provided in \cite{liu2026universal,liu2025superposition,liu2026inverse} already makes it plausible. If this view is correct, it will have implications that differ significantly from the mainstream view. Given the potential implications, it is important to explore the consequences of neural scaling universality now, even before its full verification. \textbf{We therefore argue that the current neural scaling laws have fixed exponents due to generic mechanisms, and the coefficients matter for LLM performance and design, and may be sensitive to architectural and data choices, requiring more attention and research.} We hope that a deeper exploration in the current universality class, via a closer examination of the coefficients, can help to achieve better universality classes or to identify pathways beyond it.

The rest of the paper is organized as follows. We introduce the mechanisms leading to fixed exponents with more details in Section~\ref{sec:trilogy}. We then discuss their implications, including the step-limited nature of pre-training and the importance of coefficients in determining optimal model shape and compute-optimal frontier, in Sections~\ref{sec:timevsdata}, \ref{sec:shape}, and \ref{sec:computeoptimal}, respectively. We also discuss alternative views in Section~\ref{sec:alternative} and conclude with discussions in Section~\ref{sec:discussion}.

\section{The neural scaling laws trilogy}\label{sec:trilogy}

The pre-training loss first decays as a power law in training time due to the generic properties of Softmax \cite{liu2026universal}. Here $\tau$ is the time of optimization dynamics, or dynamic time for short, related to the number of training steps via an integral over the learning rate schedule,
\begin{equation}
  \tau(t) \propto \int_0^t \eta_{t'} \mathrm{d}t',
\end{equation}
where $\tau(t)$ is the dynamic time at step $t$, and $\eta_{t'}$ is the learning rate at step $t'$. The proportionality constant may depend on the optimizer. When a model learns low-entropy or peaked distributions, Softmax becomes highly nonlinear, which produces power-law loss and gradients, and ultimately drives the loss to decay as $L \sim \tau^{-1/3}$. This is a form of universality, as the one-third exponent is insensitive to the details of the data distribution and arises from the loss function and gradient dynamics \cite{liu2026universal}.

After training to saturation, the remaining loss includes a component limited by the model's ability to represent all relevant concepts in its finite width, or embedding dimension, $m$ \cite{liu2025superposition}. In practice, LLMs encode far more features than their embedding dimensions allow, a phenomenon known as superposition \cite{arora2018linearalgebraicstructureword, elhage2022superposition}. The geometric interference among superposed representations introduces errors, leading to a residual loss $L \sim m^{-1}$. Once again, this emergent inverse width scaling of loss is robust to a range of details in the data.

Even with perfect representations, the model's transformation of those representations across $\ell$ layers can still be imperfect \cite{liu2026inverse}. Empirical analysis shows that LLM layers do not compose representations in a deep, hierarchical manner; instead, they update hidden states incrementally and similarly across layers \cite{liu2026inverse, gromov2024unreasonable,sanyal2024attention,sun2025curse,men2025shortgpt, csordas2025language}. A toy residual network analysis reveals two possible regimes, $L \sim \ell^{-3}$ if layers discretize smooth dynamics or $L \sim \ell^{-1}$ if different layers independently reduce errors by ensemble averaging. Empirical evidence supports the ensemble averaging picture, yielding the inverse-depth scaling $L \sim \ell^{-1}$ \cite{liu2026inverse}.

Combining the three scaling laws, the full loss formula takes the asymptotic form
\begin{equation}
L = \frac{c_\tau}{\tau^{1/3}} + \frac{c_m}{m} + \frac{c_\ell}{\ell} + L_0,
\end{equation}
where $L_0$ is the irreducible loss set by the entropy of the target distributions. The additive form is asymptotically accurate, i.e., expected to hold for large and well-trained models, as each term is the leading-order contribution from a distinct origin and interaction terms are treated as subleading corrections \cite{liu2026inverse,liuneural}. The coefficients $c_\tau$, $c_m$, and $c_\ell$ absorb all the data- and architecture-specific details, while the exponents are fixed by the mechanisms.

\begin{figure}
  \centering
  \includegraphics{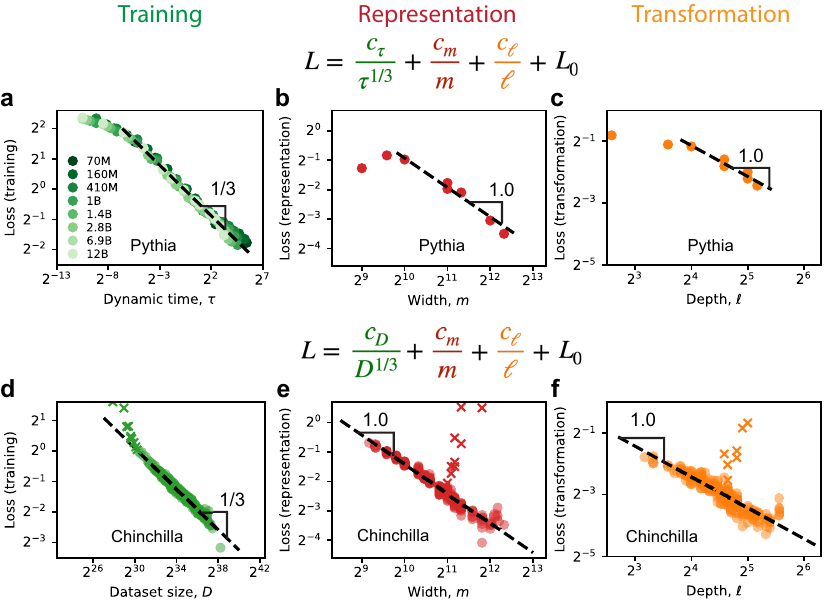}
  \caption{The neural scaling universality class, defined by one-third time scaling, inverse width scaling, and inverse depth scaling, is supported by empirical LLM results. We use the fixed-exponent formula for fitting. Dashed lines are from fitted coefficients. (a-c) The loss from Pythia models \cite{biderman2023pythia} agrees well with the scaling laws. Dashed lines span the fitting range. More details are in Appendix~\ref{app:pythia}. (d-f) With the argument that $D\propto \tau$ in online learning, the loss from Chinchilla models \cite{hoffmann2022chinchilla} also agrees well with the scaling laws with dataset size $D$ replacing dynamic time $\tau$. The crosses in the panels are ignored in fitting due to insufficient training, which also clearly do not follow the asymptotic scaling. Details are in Appendix~\ref{app:chinchilla1}.}
  \label{fig:trilogy}
\end{figure}

We next fit the loss decay with the dynamic time $\tau$, width $m$, and depth $\ell$ using Pythia models \cite{biderman2023pythia} to validate the fundamental form. We fit with fixed theoretical exponents, varying only the coefficients; previous work \cite{liu2026universal,liu2025superposition,liu2026inverse,liuneural} has shown that fitting with free exponents also works well, and we include such a comparison and all other details in Appendix~\ref{app:pythia}. For each model, we fit the loss as $L = c_\tau/\tau^{1/3} + L_{\mathrm{sat}}$ to obtain the time coefficient $c_\tau$ and the saturation loss $L_{\mathrm{sat}}$. The training part of the loss, $L - L_{\mathrm{sat}}$, is well described by the one-third time scaling across all models (Figure~\ref{fig:trilogy}a). The fact that loss from different models can collapse after subtracting $L_{\mathrm{sat}}$ fitted separately suggests that $\tau$ is the fundamental variable controlling training and that the coefficient $c_\tau$ is independent of other factors. We then fit the saturation losses across models as $L_{\mathrm{sat}} = c_m/m + c_\ell/\ell + L_0$ to obtain the width coefficient $c_m$, depth coefficient $c_\ell$, and irreducible loss $L_0$. Once $m$ and $\ell$ are large enough, the saturation loss is well described by the inverse width and depth scaling (Figure~\ref{fig:trilogy},b and c). We obtain the representation (transformation) part of loss by subtracting the fitted inverse depth (width) scaling and $L_0$ from $L_{\mathrm{sat}}$. As the Pythia suite spans a limited range of model sizes, the width and depth scaling are less strongly conclusive than the time scaling, and we provide further evidence from other model families in the following sections. Overall, the Pythia data are consistent with all three scaling laws predicted \cite{liu2026universal,liu2025superposition,liu2026inverse}, with the width and depth scaling further corroborated by the broader model families in the following sections.

\section{Pre-training may be step-limited, not data-limited}\label{sec:timevsdata}

The time scaling is fundamental yet can be transferred to the dataset size scaling emphasized in empirical neural scaling laws \cite{kaplan2020scaling,hoffmann2022chinchilla}.
When comparing models with different training dataset sizes, the last checkpoints in training curves are used. In the online training regime, each additional step consumes new tokens, so dataset size $D$ is total training steps $t$ multiplied by batch size $B$. The relation between $D$ and the corresponding dynamic time $\tau$ depends on the learning rate schedule and batch-size choice, and can be written as
\begin{equation}
  D \propto \frac{\tau}{\eta}B,
\end{equation}
where $\eta$ is the peak learning rate and the proportionality constant contains information about the shape of learning rate schedule. In the simplest case of constant $\eta$ and $B$ across models, we have $D \propto \tau$. In practice, $\eta$ and $B$ can depend on $D$ \cite{nanochat, bergsma2025power}. We want large $\eta$ to save wall time, yet it cannot be too large, as that would introduce too much noise, so $\eta\propto \sqrt{B}$ \cite{nanochat}. The batch size $B$ also increases sublinearly with $D$ \cite{bergsma2025power}. These dependencies can make $D$ scale a bit faster than $\tau$. We argue that $D\propto \tau$ is the ideal scaling for data efficiency and is approximately achieved by hyperparameter tuning when schedule shapes are comparable and $D$ (and $B$) are large. The loss scaling with time can therefore be transferred to that with dataset size, yielding
\begin{equation}
L = \frac{c_D}{D^{1/3}} + \frac{c_m}{m} + \frac{c_\ell}{\ell} + L_0.
\label{eq:trilogyD}
\end{equation}

We next fit the loss from Chinchilla models \cite{hoffmann2022chinchilla} with the dataset size scaling formula. Reconstructed loss data \cite{besiroglu2024chinchilla} are used to directly fit Equation~(\ref{eq:trilogyD}), which is expected to hold asymptotically. Data points with too small training dataset sizes (crosses in Figure~\ref{fig:trilogy}, d-f) are excluded from fitting (see Appendix~\ref{app:chinchilla1}). After fitting, we can isolate each component of the loss by subtracting the other fitted scaling terms and $L_0$ from the total loss. All three scaling laws are well supported by the LLM data (Figure~\ref{fig:trilogy}, d-f). In addition to our analysis, the dataset size scaling exponent of Chinchilla, $0.28$ reported in the original paper \cite{hoffmann2022chinchilla} and $0.37$ fitted from replication \cite{besiroglu2024chinchilla}, are both close to $1/3$. We therefore conclude that the dataset size scaling prediction agrees with LLM observations, and with more data points and a wider range of widths and depths, the width and depth scaling are better supported now.

The dataset size scaling, while empirically well established, is in our view a derived consequence of the more fundamental time scaling rather than an independent law. The mainstream framing of dataset size scaling raises the concern of eventually running out of data as models scale. Our framework instead explains the power-law loss decay through the vanishing of gradients as a power law, which follows from the nonlinear nature of Softmax. Dataset size scaling then emerges as a secondary consequence of $D \propto \tau$ in online training. Pre-training in our view is therefore step-limited rather than fundamentally limited by the amount of available unique data. Once the dataset is large enough to provide sufficient quality and diversity, further loss reduction may come from taking more optimization steps, possibly by reusing data, rather than continually increasing the amount of unique data. What ultimately matters is data quality and diversity, for accurate gradient estimation and broad downstream tasks, not just data volume.

The coefficients $c_\tau$ and $c_D$ carry direct practical significance for data efficiency. The time coefficient $c_\tau$ is the fundamental quantity governing loss decay under optimization dynamics. Once $c_\tau$ is measured, training curves under any learning rate schedule can be predicted via the corresponding dynamic time $\tau$. The dataset size coefficient $c_D$ is not fundamental in our view, and is related to $c_\tau$ through hyperparameters, making $c_D$ a lever for improving data efficiency. Understanding this relation provides a principled route to better loss per token by tuning choices like the learning rate and batch size.

\section{Coefficients determine optimal model shape}\label{sec:shape}

\begin{figure}
  \centering
  \vskip 0.2in
  \includegraphics{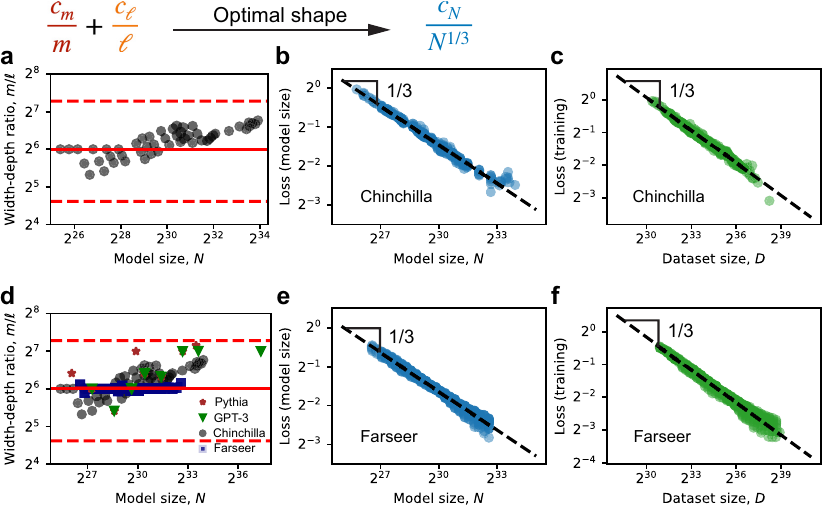}
  \caption{At the optimal shape, determined by the coefficients, width and depth scaling laws lead to a one-third model size scaling. Asymptotically, the optimal shape means $m/\ell$ is a constant. (a) Chinchilla models \cite{hoffmann2022chinchilla} are close to the optimal shape predicted (red line), where the red dashed line shows 5\% increase in loss compared to the reducible loss at the optimum. (b-c) We find that Chinchilla loss data agree well with a one-third model size scaling plus a one-third dataset size scaling. Details are in Appendix~\ref{app:chinchilla2}. (d) Different model families, i.e., Pythia \cite{biderman2023pythia}, GPT-3 \cite{brown2020gpt3}, and Farseer \cite{li2025predictable}, use aspect ratios close to our predicted optimal aspect ratio (red line and dashed lines have the same meaning as in (a)). (e-f) Loss data from Farseer models \cite{li2025predictable} agree with our scaling laws at the optimal shape as expected. Details are in Appendix~\ref{app:farseer1}.}
  \label{fig:optimalshape}
\end{figure}

The width and depth scaling laws can lead to the model size scaling law often reported in empirical neural scaling laws \cite{kaplan2020scaling,hoffmann2022chinchilla}. Since the number of parameters $N\approx 12m^2\ell$ for large $m$ and $\ell$, we minimize the loss $c_m/m + c_\ell/\ell$ at fixed $N$ and obtain the asymptotically optimal aspect ratio
\begin{equation}
  \frac{m}{\ell} = \frac{c_m}{2c_\ell}.
  \label{eq:optimalaspectratio}
\end{equation}
At this optimal shape, we have loss contributions from width and depth satisfying
\begin{equation}
  \frac{c_m}{m} \colon \frac{c_\ell}{\ell} = 2 \colon 1.
  \label{eq:optimalshapecomp}
\end{equation}
Writing $m$ and $\ell$ as a function of $N$ at the optimal shape, 
Equation~(\ref{eq:trilogyD}) yields the optimal-shape loss
\begin{equation}
  L = \frac{c_D}{D^{1/3}} + \frac{c_N}{N^{1/3}} + L_0,
  \label{eq:optimalshape}
\end{equation}
with
\begin{equation}
  c_N = 3^{4/3} c_m^{2/3} c_\ell^{1/3}.
  \label{eq:cN}
\end{equation}
We therefore have a one-third model size scaling derived from the width and depth scaling laws.

We next compare the predicted optimal aspect ratio against actual Chinchilla model shapes (Figure~\ref{fig:optimalshape}a).
With our fitted Chinchilla coefficients, the predicted optimum is $m/\ell \approx 64$ (red line), and Chinchilla models cluster near but not exactly at this value. Near the optimum, however, the loss landscape is flat, with aspect ratios from $m/\ell \approx 20$ to $180$ yielding a loss increase of less than 5\% compared to the reducible loss, i.e., $L-L_0$, at the compute-optimal point (to be introduced) with the same $N$ (red dashed lines). This is consistent with the empirical observation that shape has minor impact on loss \cite{kaplan2020scaling}, an observation possibly made near the optimum. The shape matters in principle, yet the flatness near the optimum makes precise tuning unnecessary.

Since Chinchilla models are near the optimal shape, we fit their loss with Equation~(\ref{eq:optimalshape}) (details in Appendix~\ref{app:chinchilla2}). We isolate the model size and dataset size components by subtracting the other fitted term and $L_0$ from the total loss. Both the model size component (Figure~\ref{fig:optimalshape}b) and the dataset size component (Figure~\ref{fig:optimalshape}c) follow the predicted one-third scaling well (dashed lines drawn with fitted coefficients). We therefore conclude that the one-third model size scaling around the optimal shape is supported by LLM data.

Beyond Chinchilla, we compare the predicted optimal aspect ratio against Pythia \cite{biderman2023pythia}, GPT-3 \cite{brown2020gpt3}, and Farseer \cite{li2025predictable} model families (Figure~\ref{fig:optimalshape}d). All families cluster near the predicted optimum, which is unsurprising given that practitioners likely explored aspect ratios during model design. Among them, Farseer \cite{li2025predictable} is distinctive, following $m \propto \ell$ exactly and thus falling precisely on the predicted optimal line. This exact proportionality, however, means that $c_m$ and $c_\ell$ cannot be independently identified from Farseer data due to degeneracy, yet Farseer models are well suited to testing Equation~(\ref{eq:optimalshape}).

We next fit Farseer loss with the model size plus dataset size scaling formula (details in Appendix~\ref{app:farseer1}). Isolating each component by subtracting the other term and $L_0$, we find that both the model size component (Figure~\ref{fig:optimalshape}e) and the dataset size component (Figure~\ref{fig:optimalshape}f) agree well with the one-third scaling.
The agreement across Farseer, a model family independently designed and trained, thus strengthens confidence in the generality of the three scaling laws \cite{liu2026universal,liu2025superposition,liu2026inverse} and their consequences.

Knowing $c_m$, $c_\ell$, and $c_N$ provides a principled basis for architecture design. The optimal aspect ratio (Equation~(\ref{eq:optimalaspectratio})) depends only on the ratio $c_m/c_\ell$, translating a coefficient measurement into a concrete design prescription. The model size coefficient $c_N$ (Equation~(\ref{eq:cN})), derived from $c_m$ and $c_\ell$, sets the loss floor for any given parameter budget. Since both quantities respond to architectural choices such as activation functions, attention mechanisms, and normalization schemes, as well as data composition, they are primary targets for LLM design improvements.

\section{Coefficients determine optimal token-to-parameter ratio}\label{sec:computeoptimal}

\begin{figure}
  \centering
  \includegraphics{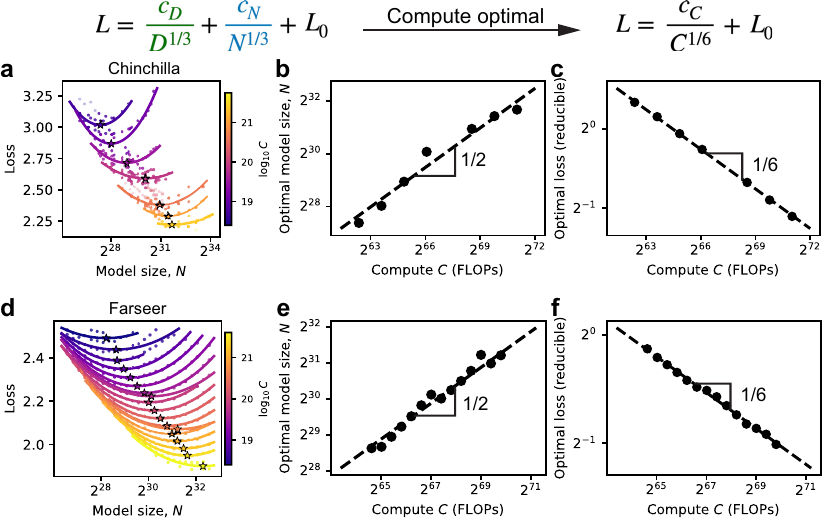}
  \caption{At the optimal token-to-parameter ratio or compute-optimal frontier, loss follows a one-sixth compute scaling. (a-c) Results for Chinchilla models \cite{hoffmann2022chinchilla}. (a) We identify isoFLOP curves and compute-optimal points (stars). (b) For the compute-optimal points, we find that the prediction $N \propto C^{1/2}$ agrees well with data. (c) The compute-optimal loss agrees well with the predicted $C^{-1/6}$ scaling. Details are in Appendix~\ref{app:chinchilla3}. (d-f) The same analysis for Farseer models \cite{li2025predictable} yields consistent results. Details are in Appendix~\ref{app:farseer2}.}
  \label{fig:computeoptimal}
\end{figure}

Building on the optimal-shape loss (Equation~(\ref{eq:optimalshape})), we next derive the compute-optimal frontier by minimizing loss at fixed compute. Since compute $C \approx 6ND$, we minimize $c_D/D^{1/3} + c_N/N^{1/3}$ subject to fixed $C$ and obtain the optimal token-to-parameter ratio
\begin{equation}
  \frac{D}{N} = \left(\frac{c_D}{c_N}\right)^3,
  \label{eq:optimalratio}
\end{equation}
which means $D \propto N$ with a proportionality constant set by the coefficients, consistent with the main argument from Chinchilla \cite{hoffmann2022chinchilla}.
At the compute-optimal frontier, we have
\begin{equation}
  \frac{c_D}{D^{1/3}} \colon \frac{c_N}{N^{1/3}} = 1 \colon 1.
\end{equation}
Combined with Equation~(\ref{eq:optimalshapecomp}), we further have the loss contributions from the three sources as
\begin{equation}
  \frac{c_D}{D^{1/3}} \colon \frac{c_m}{m} \colon \frac{c_\ell}{\ell} = 3 \colon 2 \colon 1.
\end{equation}
Representing $N$ and $D$ by $C$ at the compute-optimal frontier, the compute-optimal loss satisfies
\begin{equation}
  L = \frac{c_C}{C^{1/6}} + L_0,
  \label{eq:computeoptimal}
\end{equation}
where
\begin{equation}
  c_C = 2 \cdot 6^{1/6} c_N^{1/2} c_D^{1/2}.
  \label{eq:cC}
\end{equation}
We therefore obtain a one-sixth compute scaling of loss at the compute-optimal frontier.

We next validate the compute-optimal frontier using Chinchilla models \cite{hoffmann2022chinchilla}. We construct isoFLOP curves, i.e., loss as a function of model size $N$ at fixed compute $C \approx 6ND$, by grouping data points into bins of similar compute and fitting the loss as a function of $N$ within each bin (details in Appendix~\ref{app:chinchilla3}). The minimum of each curve gives the compute-optimal point (Figure~\ref{fig:computeoptimal}a, stars). The optimal model size grows as $N \propto C^{1/2}$ (Figure~\ref{fig:computeoptimal}b), consistent with $D \propto N \propto C^{1/2}$ from the theory prediction, and the compute-optimal loss agrees well with the predicted $C^{-1/6}$ scaling (Figure~\ref{fig:computeoptimal}c). We therefore conclude that the predicted one-sixth compute scaling at the compute-optimal frontier is well supported by Chinchilla data.

For Farseer models, we apply the same procedure of binning by compute and fitting within each bin to construct isoFLOP curves (details in Appendix~\ref{app:farseer2}) and identify compute-optimal points (Figure~\ref{fig:computeoptimal}d, stars). The optimal model size again scales as $N \propto C^{1/2}$ (Figure~\ref{fig:computeoptimal}e), and the compute-optimal loss agrees well with the $C^{-1/6}$ scaling (Figure~\ref{fig:computeoptimal}f). The agreement across Farseer further supports the predicted one-sixth compute scaling.

The coefficients $c_D$, $c_N$, and $c_C$ together govern compute allocation and efficiency. The optimal token-to-parameter ratio (Equation~(\ref{eq:optimalratio})) follows directly from $c_D$ and $c_N$, giving a compute allocation recipe grounded in the scaling laws rather than empirical rules \cite{hoffmann2022chinchilla}. The compute coefficient $c_C$ (Equation~(\ref{eq:cC})), derived from $c_N$ and $c_D$, quantifies the best loss reachable at a given compute budget. Improving either through architectural or data changes therefore offers a direct path to better compute efficiency.

\section{Alternative views}\label{sec:alternative}

\begin{figure}
  \centering
  \includegraphics{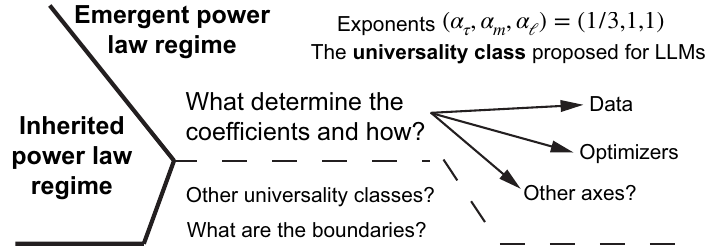}
  \caption{Neural scaling universality extends the world of scaling behaviors and provides concrete coordinates for future studies. Universality indicates that power laws in loss can emerge from generic mechanisms robust to many data details, plausibly as a consequence of strong nonlinearities such as Softmax. Alternative views attribute power-law loss to power laws in data, as in weakly nonlinear models (Section~\ref{sec:alternative}). In the current universality class, it is important to understand what determines the coefficients (Section~\ref{sec:discussion}). In the broader regime of emergent power laws, it remains unknown what other universality classes exist and what determines the boundaries between classes.}
  \label{fig:conceptual}
\end{figure}

As summarized in Figure~\ref{fig:conceptual}, the alternative view argues that the power laws in loss are inherited from power laws in data.
Intuitively, if features, skills, or tasks have power-law frequency or importance distributions and the model learns the more frequent or important ones first, the loss will decay as a power law. A wide range of frameworks formalize this intuition, including many linear toy models \cite{bahri2024explaining, bordelon2025theory, saxe2013exact} and kernel methods or effective linear models \cite{maloney2022solvable,bordelon2025feature, bordelon2021learning,lin2024scaling,bordelon2024dynamical,worschech2024analyzing,fonseca2024exactly,paquette20244+,defilippis2025scaling} trained with mean squared error (MSE). In these frameworks, the loss exponent is jointly determined by the data power-law exponent and architectural parameters such as the number of linear layers \cite{bordelon2025feature,bordelon2025theory, saxe2013exact}. Predicting the exponents in practice therefore requires measuring the power-law structure of language, while the connection between idealized toy setups, such as linear regression over power-law inputs, and real LLM training has not been well established. The implied research agenda is to characterize the power-law structure of training data, especially in real language, and to understand how architecture translates data exponents into loss exponents.

With the same intuition, some frameworks with high-level arguments like fitting data manifold predict that loss exponents depend only on data structure, independent of architecture \cite{sharma2022neural,michaud2023quantization,arora2023theory,liu2025physicsskilllearning,cagnetta2026deriving}. In this view, the optimal model shape and compute-optimal token-to-parameter ratio depend sensitively on the measured data statistics and may not reduce to the simple coefficient-driven expressions predicted here (Equations~(\ref{eq:optimalaspectratio}) and (\ref{eq:optimalratio})). The resulting optimal relationships may therefore vary across domains and data compositions, requiring separate characterization for each setting.

Alternative proposals, together with the three papers \cite{liu2026universal,liu2025superposition,liu2026inverse}, constitute a zoo of different behaviors and theories, each plausible in its own regime. Language statistics predict dataset scaling exponent $\alpha_D$ to be $0.14\sim 0.19$ \cite{cagnetta2026deriving}, which agrees with early LLM training. However, later training follows a steeper power law with exponents around $1/3$ \cite{liu2026universal}, which is closer to $0.28$ reported in Chinchilla \cite{hoffmann2022chinchilla}. One possible reconciliation is that these theories describe different training regimes. In the early stage, LLMs are in the high-temperature regime, with relatively flat output distributions where nonlinearity in loss is weak, leading to data-sensitive exponents \cite{liu2026universal}. Later, in the low-temperature regime, where output distributions become peaked and Softmax becomes strongly nonlinear, loss follows the one-third time scaling. Both the data-sensitive and neural scaling universality regimes exist, each possibly containing more classes of behaviors than those discussed. The key questions for the field are which regime is most relevant to language processing, and what conditions determine which regime applies.

\section{Discussion}\label{sec:discussion}

\begin{figure}
  \centering
  \includegraphics{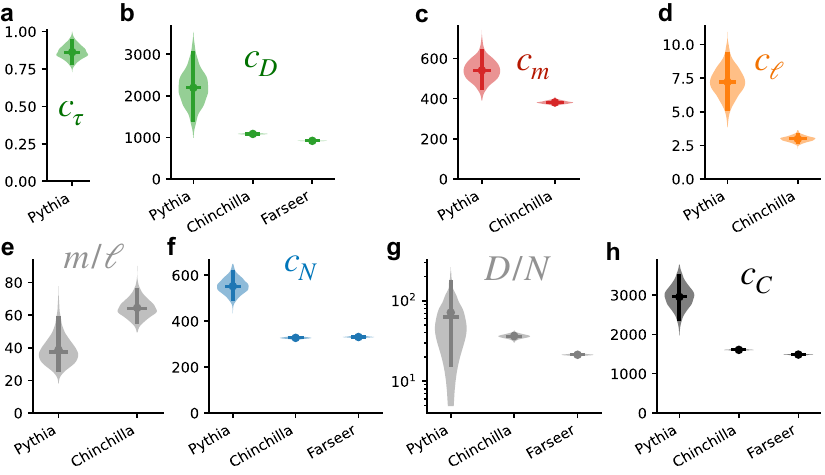}
  \caption{Coefficients and optimal ratios inferred from fitting vary across model families. (a) Time coefficient $c_\tau$. (b) Dataset size coefficient $c_D$. (c) Width coefficient $c_m$. (d) Depth coefficient $c_\ell$. (e) Optimal aspect ratio $m/\ell$ predicted from coefficients. (f) Model size coefficient $c_N$ at the optimal shape. (g) Optimal token-to-parameter ratio $D/N$ predicted from coefficients. (h) Compute coefficient $c_C$ at the compute-optimal frontier. The violin plots show distributions obtained from bootstrap analysis. Within each violin, the horizontal line indicates the median, the vertical line indicates the 95\% confidence interval, and the dot indicates the point estimation. Details are in Appendices~\ref{app:pythia}, \ref{app:chinchilla1}, and \ref{app:farseer1}.}
  \label{fig:discussion}
\end{figure}

To conclude, the three papers \cite{liu2026universal,liu2025superposition,liu2026inverse}, which identify generic mechanisms for the one-third time scaling, inverse width scaling, and inverse depth scaling, suggest that the universality regime is most relevant to current LLMs, making it timely to explore its implications now. We therefore propose \textbf{neural scaling universality, a hypothesis that different universality classes of neural scaling laws exist, each defined by a set of fixed exponents and each containing a wide range of architectures and data}, with the three papers forming one universality class in our terminology. This view differs from and complements the mainstream one, which requires power-law structure in data to explain power-law loss and predicts sensitive dependence of loss exponents on data power laws, forming a larger zoo of scaling behaviors in neural networks. Within this universality class, the scaling coefficients depend on data, architecture, and hyperparameters, and determine the optimal model shape and the compute-optimal frontier. \textbf{We therefore argue that studying the coefficients of scaling laws is the key near-term direction for performance improvement, and that a detailed exploration of the current universality class may point the way to better ones.}

The three scaling laws with time, width, and depth are the most fundamental, as are their coefficients $c_\tau$, $c_m$, and $c_\ell$. For Pythia models \cite{biderman2023pythia}, we fit the three fundamental coefficients (Figure~\ref{fig:trilogy}, a-c) and derive all other quantities (Figure~\ref{fig:discussion}, b, e-h). For Chinchilla models \cite{hoffmann2022chinchilla}, we fit $c_D$, $c_m$, and $c_\ell$ (Figure~\ref{fig:trilogy}, d-f) and derive the rest (Figure~\ref{fig:discussion}, e-h). For Farseer models \cite{li2025predictable}, we fit $c_D$ and $c_N$ (Figure~\ref{fig:optimalshape}, e and f) and derive the rest (Figure~\ref{fig:discussion}, g and h). These coefficients differ across model families, which is important to understand, yet the variation is modest given the large uncertainties.

Future work should characterize the quantitative dependence of the coefficients on data, architectures, optimizers, and hyperparameters. Concretely, this means measuring $c_\tau$, $c_m$, and $c_\ell$ under controlled changes to data composition and architecture, and measuring how $c_D$ shifts with learning rate, batch size, schedule shape, and optimizer. The time coefficient $c_\tau$ depends on output distribution statistics \cite{liu2026universal}, though the precise mechanism remains unclear; $c_m$ is affected by feature frequency distributions \cite{liu2025superposition}; and architectural choices such as activation function, position embedding, and layer normalization can change single-layer expressivity and therefore $c_\ell$. Since $c_D$ is related to $c_\tau$ through hyperparameters (Section~\ref{sec:timevsdata}), tuning learning rate, batch size, schedule, and optimizer may be one of the most accessible routes to improving data and compute efficiency. Consistent with this view, new optimizers like Muon appear to provide a constant-factor speedup compared with Adam while leaving the scaling exponent similar \cite{liu2025muon}. The larger $c_D$ (Figure~\ref{fig:discussion}b) and optimal $D/N$ (Figure~\ref{fig:discussion}g) of Pythia models may be partly due to the large constant batch size used. Since $c_N$ and $c_C$ are derived from these coefficients (Sections~\ref{sec:shape} and \ref{sec:computeoptimal}), future studies need only characterize the three fundamental coefficients and the relation between $c_D$ and $c_\tau$ to complete this theory.

Beyond the dense Transformer models on which neural scaling laws have been mostly studied, the framework also predicts how architectural modifications may enter. Mixture of experts \cite{fedus2022switch, clark2022unified}, looped transformers \cite{fan2024looped, prairie2026parcae}, and attention residual networks \cite{team2026attention} should not alter the three scaling laws \cite{liu2026universal,liu2025superposition,liu2026inverse} if the underlying mechanisms are unchanged, but may change the coefficients, especially $c_\ell$, which governs transformation. Architectural parameters such as the number of experts or loop iterations may therefore introduce extra tuning axes, while also modifying parameter count, compute budget, optimal shape, and the compute-optimal frontier. Taking the three scaling laws as a starting point reduces the question to how the coefficients change, from which other consequences follow.

Understanding the current universality class may better characterize its boundaries and indicate how to move to better classes. Preliminary evidence indeed suggests the existence of other classes \cite{chen2026superposition}, which is encouraging.

We hope that these arguments will inspire more studies on neural scaling universality, revealing deeper understanding, proving or disproving our hypothesis, and ultimately helping to achieve more efficient and more interpretable LLMs.

\begin{ack}
The authors are grateful for feedback and suggestions from Toni Liu and Amil Dravid.
\end{ack}

\bibliographystyle{unsrt}
\bibliography{refs}


\appendix

\section{Pythia models}\label{app:pythia}

We fit the scaling law
\begin{equation}
  L = \frac{c_\tau}{\tau^{1/3}} + \frac{c_m}{m} + \frac{c_\ell}{\ell} + L_0
\end{equation}
to loss data of the Pythia suite \cite{biderman2023pythia}, which contains eight publicly available language models of sizes $70$M, $160$M, $410$M, $1$B, $1.4$B, $2.8$B, $6.9$B, and $12$B, all trained for $143{,}000$ steps on the Pile \cite{gao2020pile} with batch size $B = 2\times10^6$ tokens. For each model, we evaluate the loss at $13$ checkpoints (steps $128$ to $143{,}000$), loading model weights from HuggingFace and evaluating the per-token cross-entropy loss on $4{,}800$ documents ($\sim2$M tokens) drawn from FineWeb \cite{penedo2024fineweb} (a held-out corpus distinct from the training set). Model widths and depths are read from the public Pythia model configurations on HuggingFace, and parameter counts are taken from the Pythia model size labels.
We also record the standard deviation of the output logits as an auxiliary quantity tracking the sharpening of next-token predictions over training.

The dynamic time is computed as $\tau = \sum_{t'=1}^{t}\eta_{t'}$, where $\eta_{t'}$ is the per-step learning rate under Pythia's cosine schedule (linear warmup over the first 1\% of training, then cosine decay to 10\% of peak).
To be rigorous, as noted in the main text, $\tau \propto \sum_{t'=1}^{t}\eta_{t'}$. The proportionality constant depends on the optimizer and possibly also hyperparameters such as momentum and batch size. The Adam optimizer may effectively follow gradient descent via zigzagging \cite{liu2026universal}. In that case, learning rate of Adam is related to the learning rate of the effective gradient descent dynamics, yet the proportionality constant is affected by zigzagging behaviors, gradient noise, and therefore momentum and batch size. Since Pythia models use the same optimizer, momentum, and a large batch size across all models, we expect the proportionality constant to be similar across models, and absorb it into $c_\tau$ by directly using $\tau = \sum_{t'=1}^{t}\eta_{t'}$.
The relationship between $\tau$, the number of training tokens $D$, and the peak learning rate $\eta$ is approximately $\tau \approx (\eta/2B)\,D$, where $1/2$ is from the integral of the cosine function. Then, the time coefficient can be converted to a data coefficient via
\begin{equation}
  c_D = c_\tau \left(\frac{\eta}{2B}\right)^{-1/3}.
\end{equation}

\paragraph{Fitting $c_\tau$.}
For each model size $k$ we fix the time exponent at $1/3$ and fit $c_{\tau,k}$ and the saturation loss $L_{\mathrm{sat},k}$ by nonlinear least squares to
$$
\sum_t[\log L_t - \log(c_{\tau,k}/\tau_t^{1/3} + L_{\mathrm{sat},k})]^2,
$$
excluding the first one or two checkpoints to avoid the warmup transient.
The eight per-model estimates $\{c_{\tau,k}\}$ are treated as independent measurements of a universal constant; we report the mean $c_\tau$ and standard deviation $\sigma_{c_\tau}$ across models. Results are shown in Figure~\ref{fig:trilogy}a.
As supporting evidence that the fixed exponent $1/3$ is justified, fitting with a free exponent across four representative models (70M, 410M, 2.8B, 12B) yields time scaling exponent $\alpha_\tau \in [0.31, 0.37]$, all consistent with $1/3$.

\paragraph{Fitting $c_m$ and $c_\ell$.}
The saturation losses $\{L_{\mathrm{sat},k}\}$ reflect architectural capacity.
We jointly fit the six largest models (410M--12B; the two smallest are excluded because their saturation losses are not yet dominated by the architectural limits) by nonlinear least squares to
\begin{equation}
  L_{\mathrm{sat}} = \frac{c_m}{m} + \frac{c_\ell}{\ell} + L_0,
\end{equation}
obtaining $c_m$, $c_\ell$, and the irreducible loss $L_0$.
Standard errors $\sigma_{c_m}$ and $\sigma_{c_\ell}$ are taken from the square root of the diagonal of the covariance matrix returned by the solver.
Results are shown in Figure~\ref{fig:trilogy}, b and c.

\paragraph{Derived scaling constants.}
Given $N \approx 12m^2\ell$, the optimal width-to-depth ratio at fixed $N$ satisfies $m/\ell = c_m/(2c_\ell)$, giving an effective model size coefficient $c_N = 3^{4/3}c_m^{2/3}c_\ell^{1/3}$ and compute coefficient $c_C = 2\cdot6^{1/6}c_N^{1/2}c_D^{1/2}$.
The compute-optimal token-to-parameter ratio is $D/N = (c_D/c_N)^3$.

\paragraph{Uncertainty quantification.}
We propagate uncertainties via a parametric bootstrap with $10^3$ draws.
The architectural constants $(c_m, c_\ell)$ are drawn jointly from the bivariate normal implied by the least-squares covariance, preserving their correlation.
The data constant $c_D$ is drawn independently as a normal random variable with the mean $c_D$ and standard deviation $\sigma_{c_D}$ estimated across the eight per-model fits.
Each draw is propagated through the nonlinear expressions above; we report mean and standard deviation from the resulting empirical distributions.

The fitted and derived constants are summarized in Table~\ref{tab:pythia}, and the derived quantities are shown in Figure~\ref{fig:discussion}, b, e-h.

\begin{table}
\centering
\caption{Scaling constants fitted from Pythia models. Uncertainties for $c_\tau$ and $c_D$ are standard deviations across the eight per-model fits; uncertainties for $c_m$ and $c_\ell$ are standard errors from the nonlinear least-squares solver; uncertainties for derived quantities are propagated via parametric bootstrap.}
\label{tab:pythia}
\begin{tabular}{lll}
\toprule
Quantity & Description & Value \\
\midrule
$c_\tau$       & Time coefficient       & $0.865 \pm 0.045$ \\
$c_D$          & Dataset size coefficient & $2178 \pm 421$ \\
$c_m$          & Width coefficient      & $539 \pm 52$ \\
$c_\ell$       & Depth coefficient      & $7.26 \pm 1.11$ \\
\midrule
$c_N$          & Model size coefficient & $555 \pm 46$ \\
$c_C$          & Compute coefficient    & $2964 \pm 311$ \\
$m/\ell$       & Optimal aspect ratio   & $37.2 \pm 6.7$ \\
$D/N$          & Optimal token-to-parameter ratio & $60 \pm 38$ \\
\bottomrule
\end{tabular}
\end{table}

\section{Chinchilla models}

\subsection{Data, width, and depth}\label{app:chinchilla1}

We fit Equation~(\ref{eq:trilogyD}) to $245$ data points reconstructed from Chinchilla scaling-law experiments \cite{besiroglu2024chinchilla}, spanning a wide range of model widths $m$, depths $\ell$, and training dataset sizes $D$. Widths, depths, parameter counts, and training dataset sizes are taken from the Chinchilla paper \cite{hoffmann2022chinchilla}.

\paragraph{Fitting.}
Unlike the Pythia analysis, where the time and saturation terms are fitted in separate stages, all four parameters $(c_D,\,c_m,\,c_\ell,\,L_0)$ are fitted jointly here by nonlinear least squares to 
$$
\sum_i[\log L_i - \log(c_D/D_i^{1/3} + c_m/m_i + c_\ell/\ell_i + L_0)]^2,
$$
using $50$ random initializations and retaining the best solution.
Data points with the smallest $D$ are excluded because the proportionality $D \propto \tau$ (Section~\ref{sec:timevsdata}) breaks down at insufficient training, so those early-training points do not yet lie on the asymptotic power law.
To select the cutoff, we scan $k = 0, 5, \ldots, 40$ (number of smallest-$D$ points removed) and choose the $k$ that minimizes the mean of the standard errors of the four fitted parameters, where each standard error is taken from the square root of the corresponding diagonal entry of the covariance matrix returned by the solver for the log-residual least-squares problem. Too few removals retain early-training points that violate the power law and inflate the error; too many discard valid asymptotic data and also inflate it.
We observed that $k = 10$ minimizes the uncertainty and chose $k = 10$ for further analysis, leaving $235$ points; the excluded points are shown as crosses in Figure~\ref{fig:trilogy}, d-f.
As supporting evidence that the fixed exponents are justified, fitting with free exponents on the same $235$ points yields width exponent $\alpha_m = 0.87 \pm 0.06$, depth exponent $\alpha_\ell = 0.97 \pm 0.20$, and dataset exponent $\alpha_D = 0.34 \pm 0.01$, broadly consistent with $1$, $1$, and $1/3$, respectively.

\paragraph{Derived scaling constants.}
From the fitted $(c_D, c_m, c_\ell)$, we derive the optimal aspect ratio, model size coefficient, compute coefficient, and optimal token-to-parameter ratio via Equations~(\ref{eq:optimalaspectratio}), (\ref{eq:cN}), (\ref{eq:cC}), and~(\ref{eq:optimalratio}), respectively.

\paragraph{Uncertainty quantification.}
We propagate uncertainties via a bootstrap with $10^3$ draws.
In each draw, the $235$ retained data points are resampled with replacement and the full four-parameter fit is repeated. For each draw, the corresponding derived quantities are computed from the fitted parameters, which therefore also have a distribution across draws.
We report bootstrap mean and standard deviation across draws; the distributions are shown in Figure~\ref{fig:discussion}.

The fitted and derived constants are summarized in Table~\ref{tab:chinchilla1}.

\begin{table}
\centering
\caption{Scaling constants fitted from Chinchilla models. All uncertainties are bootstrap standard deviations from $10^3$ draws.}
\label{tab:chinchilla1}
\begin{tabular}{lll}
\toprule
Quantity & Description & Value \\
\midrule
$c_D$          & Dataset size coefficient & $1084 \pm 11$ \\
$c_m$          & Width coefficient      & $381 \pm 8$ \\
$c_\ell$       & Depth coefficient      & $2.98 \pm 0.21$ \\
$L_0$          & Irreducible loss       & $1.785 \pm 0.006$ \\
\midrule
$c_N$          & Model size coefficient & $327 \pm 4$ \\
$c_C$          & Compute coefficient    & $1605 \pm 10$ \\
$m/\ell$       & Optimal aspect ratio   & $64.4 \pm 5.8$ \\
$D/N$          & Optimal token-to-parameter ratio & $36.4 \pm 2.1$ \\
\bottomrule
\end{tabular}
\end{table}

\subsection{Data and model size}\label{app:chinchilla2}

We fit Equation~(\ref{eq:optimalshape}) to the same $245$ data points, using the same joint nonlinear least squares procedure and $k = 10$ cutoff as in Section~\ref{app:chinchilla1}, leaving $235$ points.
The fit has three parameters $(c_D, c_N, L_0)$. Since Chinchilla models are near the optimal shape (Figure~\ref{fig:optimalshape}a), $N$ collapses width and depth into a single variable.
The fitted results are shown in Figure~\ref{fig:optimalshape}, b-c, and summarized in Table~\ref{tab:chinchilla2}.

\begin{table}
\centering
\caption{Scaling constants fitted from Chinchilla models using the optimal-shape form (Equation~(\ref{eq:optimalshape})). Uncertainties are standard errors from the nonlinear least-squares solver.}
\label{tab:chinchilla2}
\begin{tabular}{lll}
\toprule
Quantity & Description & Value \\
\midrule
$c_D$          & Dataset size coefficient & $1076 \pm 8$ \\
$c_N$          & Model size coefficient & $374 \pm 2$ \\
$L_0$          & Irreducible loss       & $1.762 \pm 0.003$ \\
\bottomrule
\end{tabular}
\end{table}

\subsection{Optimal compute analysis}\label{app:chinchilla3}

We identify compute-optimal points from isoFLOP profiles using the same $235$ data points as in Section~\ref{app:chinchilla1}, with compute estimated as $C = 6ND$.
We divide the data into $10$ equal-width bins in $\log_{10} C$ over the range $\log_{10} C \in [18.4, 21.75]$, assigning each point to the nearest bin center.
Bins with fewer than $6$ points are excluded.

\paragraph{IsoFLOP profile fitting.}
Within each bin, $C$ is approximately fixed.
Substituting $D = C/(6N)$ into Equation~(\ref{eq:optimalshape}) gives the isoFLOP profile
\begin{equation}
  L = a\,N^{-1/3} + b\,N^{1/3} + L_0,  \label{eq:isoflop}
\end{equation}
where $a = c_N$ and $b = c_D(6/C)^{1/3}$ depend on the fixed-exponent coefficients and the bin's compute $C$.
We fit $a$, $b$, and $L_0$ jointly by nonlinear least squares with multiple random initializations, retaining the best solution.
Setting $\partial L/\partial N = 0$ in Equation~(\ref{eq:isoflop}) yields the compute-optimal model size
\begin{equation}
  N^* = \left(\frac{a}{b}\right)^{3/2},
\end{equation}
at which the compute-optimal loss is
\begin{equation}
  L^* = 2\sqrt{ab} + L_0.
\end{equation}
To guard against degenerate fits where the minimum lies far outside the observed data range, we retain only bins where $N^*$ falls within a factor of $5$ of the observed $N$ range of that bin.
This yields $7$ valid compute-optimal points, shown as stars in Figure~\ref{fig:computeoptimal}a.

\paragraph{Results.}
Power-law fits to $N^*$ and $D^* = C/(6N^*)$ as functions of $C$ yield $N^* \propto C^{0.51}$ and $D^* \propto C^{0.49}$ (Figure~\ref{fig:computeoptimal}b), both consistent with the theoretical $C^{1/2}$ scaling.
Fitting the compute-optimal loss $L^*$ with Equation~(\ref{eq:computeoptimal}) at the fixed exponent $1/6$ gives $c_C = 1710$ and $L_0 = 1.76$, consistent with the data ($R^2 = 0.9995$; Figure~\ref{fig:computeoptimal}c).

\section{Farseer models}\label{app:farseer}

\subsection{Data and model size}\label{app:farseer1}

We fit Equation~(\ref{eq:optimalshape}) to Farseer loss data \cite{li2025predictable} comprising $404$ data points spanning a range of model sizes $N$ and training dataset sizes $D$. Model sizes and training dataset sizes are taken from the Farseer code repository associated with \cite{li2025predictable}.
Because Farseer follows $m \propto \ell$ exactly, the individual width and depth coefficients $c_m$ and $c_\ell$ are degenerate and cannot be identified separately; only the combined model size coefficient $c_N$ in Equation~(\ref{eq:optimalshape}) is identifiable.

\paragraph{Fitting.}
All three parameters $(c_D, c_N, L_0)$ are fitted jointly by nonlinear least squares to
$$
\sum_i [\log L_i - \log(c_N/N_i^{1/3} + c_D/D_i^{1/3} + L_0)]^2,
$$
using $50$ random initializations and retaining the best solution.
Data points with the highest loss are excluded because they correspond to under-trained runs where the asymptotic power law does not yet apply.
To select the cutoff, we scan $k = 0, 20, \ldots, 160$ (number of highest-loss points removed) and choose the $k$ that minimizes the mean of the standard errors of the three fitted parameters, where each standard error is the square root of the corresponding diagonal entry of the solver covariance matrix for the log-residual least-squares problem.
We select $k = 80$, leaving $324$ points.
As supporting evidence that the fixed exponents are justified, fitting a free-exponent form $c_N N^{-\alpha_N} + c_D D^{-\alpha_D} + L_0$ on the same $324$ points yields $\alpha_N = 0.30 \pm 0.01$ and $\alpha_D = 0.349 \pm 0.009$, both broadly consistent with the theoretical $1/3$.
The fitted results are shown in Figure~\ref{fig:optimalshape}, e-f.

\paragraph{Derived scaling constants.}
From the fitted $(c_D, c_N)$, we derive the compute coefficient and optimal token-to-parameter ratio via Equations~(\ref{eq:cC}) and~(\ref{eq:optimalratio}), respectively.

\paragraph{Uncertainty quantification.}
We propagate uncertainties via a bootstrap with $10^3$ draws.
In each draw, the $324$ retained data points are resampled with replacement and the full three-parameter fit is repeated; for each draw, the derived quantities are computed from the fitted parameters.
Of the $10^3$ resamples, $926$ converged to nondegenerate solutions, defined as finite fits with positive coefficients, and are retained.
We report bootstrap mean and standard deviation across these draws; the distributions are shown in Figure~\ref{fig:discussion}, g and h.

The fitted and derived constants are summarized in Table~\ref{tab:farseer1}.

\begin{table}
\centering
\caption{Scaling constants fitted from Farseer models using the optimal-shape form (Equation~(\ref{eq:optimalshape})). All uncertainties are bootstrap standard deviations from $10^3$ draws.}
\label{tab:farseer1}
\begin{tabular}{lll}
\toprule
Quantity & Description & Value \\
\midrule
$c_D$          & Dataset size coefficient & $918 \pm 6$ \\
$c_N$          & Model size coefficient & $331 \pm 3$ \\
$L_0$          & Irreducible loss       & $1.542 \pm 0.005$ \\
\midrule
$c_C$          & Compute coefficient    & $1486 \pm 10$ \\
$D/N$          & Optimal token-to-parameter ratio & $21.3 \pm 0.5$ \\
\bottomrule
\end{tabular}
\end{table}

\subsection{Optimal compute analysis}\label{app:farseer2}

We identify compute-optimal points from isoFLOP profiles following the same procedure as in Section~\ref{app:chinchilla3}. An isoFLOP profile is the loss as a function of model size $N$ at approximately fixed compute $C \approx 6ND$; its minimum defines the compute-optimal model size $N^*$, the corresponding dataset size $D^* = C/(6N^*)$, and the compute-optimal loss $L^*$.
We apply the same exclusion of the $80$ highest-loss points as in Section~\ref{app:farseer1} and further restrict to data with $\log_{10} C \in [18.40, 21.61]$, yielding $323$ points.
We divide the data into equal-width bins in $\log_{10} C$, merging adjacent bins until every bin contains at least $10$ points, yielding $18$ final bins.
Within each bin, we fit the isoFLOP profile form (Equation~(\ref{eq:isoflop})) by nonlinear least squares and locate $N^*$, $D^*$, and $L^*$ from the fitted minimum following Section~\ref{app:chinchilla3}.
All $18$ bins pass the validity check ($N^*$ falls within a factor of $5$ of the observed $N$ range), giving $18$ compute-optimal points shown as stars in Figure~\ref{fig:computeoptimal}d.

\paragraph{Results.}
Power-law fits to $N^*$ and $D^*$ as functions of $C$ yield $N^* \propto C^{0.52}$ and $D^* \propto C^{0.48}$ (Figure~\ref{fig:computeoptimal}e), both consistent with the theoretical $C^{1/2}$ scaling.
Fitting the compute-optimal loss $L^*$ with Equation~(\ref{eq:computeoptimal}) at the fixed exponent $1/6$ gives $c_C = 1590$ and $L_0 = 1.52$, consistent with the data ($R^2 = 0.9953$; Figure~\ref{fig:computeoptimal}f).



\end{document}